\documentclass[journal]{IEEEtran}

\usepackage{cite}
\usepackage{amsmath,amssymb,amsfonts}
\usepackage{algorithm,algorithmic}
\usepackage{graphicx}
\usepackage{booktabs}
\usepackage{array}
\usepackage{multirow}
\usepackage{hhline}
\usepackage{textcomp}
\usepackage{fancyvrb}
\usepackage{url}
\usepackage[hidelinks]{hyperref}
\graphicspath{{./}}
\title{Bio-MF: Low-Latency and High-Fidelity EEG-to-fNIRS Cross-Modal Generation for Hybrid Motor-Imagery Brain--Computer Interfaces}

\author{Boyuan~Zhao, Sifan~Zhang, and Luping~Chen%
\thanks{Boyuan Zhao and Luping Chen are with the Key Laboratory of Modern Teaching Technology, Shaanxi Normal University, Xi'an 710062, China (e-mail: 20243259@snnu.edu.cn; 2023301681@snnu.edu.cn).}%
\thanks{Sifan Zhang is with Microsoft, Beijing, China (e-mail: sifanzhang@microsoft.com).}%
\thanks{Corresponding author: Boyuan Zhao.}}

\begin{document}

\maketitle

\begin{abstract}
Hybrid motor-imagery brain-computer interfaces (MI-BCIs) combining EEG and fNIRS can outperform EEG-only systems by exploiting complementary electrophysiological and hemodynamic information. To obtain such hybrid information when paired EEG-fNIRS acquisition is unavailable or inconvenient, recent studies have focused on EEG-to-fNIRS cross-modal generation. However, existing methods still suffer from slow generation and often require pretraining, limiting their use in hybrid MI-BCI scenarios. Although one-step generative models offer an attractive route to low-latency synthesis, removing the iterative refinement process can reduce generation fidelity and introduce non-physiological artifacts. To address these problems, this paper proposes Bio-MF, a latent-free one-step MeanFlow framework for EEG-conditioned fNIRS generation. Bio-MF directly predicts clean fNIRS signals rather than noise in the raw fNIRS space, converts this clean-signal output into MeanFlow velocity supervision, and completes inference with one network evaluation. To preserve task-relevant hemodynamic structure under heterogeneous sensor layouts, Bio-MF integrates Spatial-Temporal Interactive 4D Encoding, cross-modal classifier-free guidance, and noise-level-gated FFT regularization. On Dataset 1, EEG + synthetic fNIRS improves ACC over EEG-only by 3.37 and 4.15 percentage points for HbR and HbO, respectively. On Dataset 2, the corresponding gains remain 2.98 and 2.50 percentage points under the unseen 64-channel EEG montage. On an RTX PRO 6000 GPU, Bio-MF generates one fNIRS trial in 7.0 ms, corresponding to an 857x speedup over the 1000-step SCDM latency. These results show that Bio-MF enables fast EEG-to-fNIRS synthesis while preserving task-relevant generation quality for hybrid MI-BCIs. Our code is available at \url{https://github.com/psychosiwa/Bio-MF}.
\end{abstract}

\begin{IEEEkeywords}
Brain-computer interface, EEG-to-fNIRS generation, MeanFlow, multimodal neuroimaging, motor imagery, single-step generation
\end{IEEEkeywords}

\IEEEpeerreviewmaketitle

\section{Introduction}

Motor imagery brain-computer interfaces (MI-BCIs) decode sensorimotor activity induced by imagined limb movements and provide a non-muscular control pathway for rehabilitation, assistive control, and human-machine interaction [1]--[3]. Electroencephalography (EEG) is widely used in MI-BCIs because it is non-invasive, portable, relatively inexpensive, and has millisecond-level temporal resolution. Most EEG-based MI-BCI methods therefore focus on extracting discriminative sensorimotor rhythm features from EEG signals, using spatial filtering, time-frequency analysis, or neural classifiers to decode left- and right-hand motor imagery [4]--[7].

Despite these advantages, EEG-only MI decoding remains challenging. EEG signals have a low signal-to-noise ratio and are affected by ocular artifacts, muscle activity, volume conduction, inter-subject variability, and session-to-session non-stationarity [4]--[7]. These factors are particularly relevant for single-trial MI decoding, where event-related desynchronization and synchronization patterns can be weak and variable across users and recording sessions. In addition, EEG has limited spatial localization, which restricts its ability to characterize local cortical activation associated with motor imagery. These limitations have motivated multimodal BCI studies that combine EEG with complementary neural or physiological signals.

\begin{sloppypar}
Functional near-infrared spectroscopy (fNIRS) is a complementary modality. It records task-related changes in oxygenated and deoxygenated hemoglobin [8]--[11]. Although fNIRS has lower temporal resolution than EEG because of the delayed hemodynamic response, it can provide spatial and metabolic information related to cortical activation. Hybrid EEG-fNIRS systems can therefore use EEG to capture fast electrophysiological dynamics and fNIRS to provide complementary hemodynamic information. Previous studies have reported that EEG-fNIRS fusion can improve MI decoding performance and stability compared with EEG-only systems in several settings [12]--[18].
\end{sloppypar}

However, simultaneous EEG-fNIRS acquisition is less convenient than EEG-only recording. EEG electrodes and fNIRS sources/detectors must share the scalp surface, which leads to sensor-layout conflicts and increases preparation time, optical coupling requirements, and sensitivity to hair obstruction and source-detector spacing [19]--[23]. The hybrid setup can also be inconvenient for repeated calibration, portable BCI use, and feedback-based settings where preparation time and user comfort are important. These constraints motivate EEG-conditioned synthesis of task-related fNIRS signals for hybrid MI decoding when only EEG is available.

Cross-modal generation has also been extended to neural-signal synthesis, including EEG-to-fMRI generation [24]. For EEG-to-fNIRS synthesis, SCDM models spatial and multi-scale temporal relations [25], whereas TADM combines unified pretraining with latent diffusion to improve transfer across tasks and sensor configurations [26]. These studies establish EEG-conditioned generation as a route to auxiliary representations for BCI decoding.

Despite these advances, existing EEG-to-fNIRS generation methods still have limitations for latency-sensitive hybrid MI-BCI use. SCDM uses an iterative diffusion paradigm with a U-Net-like denoising network, spatial cross-modal generation modules, multi-scale temporal representation modules, and precomputed correlation matrices to guide EEG-to-fNIRS mapping [25]. This design supports EEG-to-fNIRS synthesis, but diffusion inference requires a long reverse denoising chain, which increases generation latency. In addition, precomputed correlation matrices and projection-based spatial representations may make the model dependent on a specific sensor layout, limiting adaptation to different EEG montages or fNIRS geometries. TADM reduces part of the modeling burden by using unified pretraining and latent diffusion [26]. However, unified pretraining can increase training and deployment cost, while VAE latent compression introduces an additional encoder-decoder reconstruction path, may attenuate subtle HbO/HbR variations, and still retains iterative sampling latency because generation is performed in a latent diffusion process. Moreover, convolutional U-Net architectures are effective for regular-grid signals but usually require input channels to be arranged on a fixed grid or mapped through predefined spatial transformations. Such limitations can make the model dependent on a specific montage and less flexible when the number or physical placement of sensors changes. In contrast, a token-based Transformer formulation can represent EEG and fNIRS patches together with their physical coordinates and auxiliary controls in a shared sequence, allowing the model to process heterogeneous sampling rates and non-uniform sensor layouts more naturally [32]--[34].

To address the above issues, we propose Bio-MF, a Transformer-based latent-free single-step MeanFlow framework for EEG-conditioned fNIRS generation in hybrid MI-BCIs. Bio-MF replaces multi-step diffusion sampling with a flow-matching one-step generation paradigm. MeanFlow learns an interval-average velocity field that enables 1-NFE generation without progressive distillation or a long reverse denoising trajectory [30], while Pixel Mean Flows show that latent-free signal-space outputs can be combined with velocity-space supervision [31]. Following this idea, Bio-MF directly predicts the clean fNIRS signal rather than a noise residual in the raw fNIRS signal space and converts this clean-signal output into MeanFlow velocity supervision. This avoids VAE compression and reconstruction, keeps the generated output in the physiological signal domain, and removes both latent decoding and iterative denoising from the inference path. Because the network output remains an fNIRS signal, physiological and spectral constraints can be applied directly to the generated content, making it possible to optimize signal quality while retaining one-step inference speed. To further mitigate the fidelity loss caused by removing iterative refinement, Bio-MF uses a two-stage noise-level-gated FFT regularizer that applies spectral alignment only in low-noise states, thereby reducing frequency-domain artifacts while avoiding conflicts with MeanFlow velocity learning.

Bio-MF further combines geometry-aware token encoding with controllable cross-modal guidance to improve EEG-to-fNIRS generation under heterogeneous sensor layouts. First, inspired by REVE [36], Bio-MF represents EEG electrodes and fNIRS channels using continuous four-dimensional coordinates $(x,y,z,t)$, where $(x,y,z)$ encodes the physical sensor location and $t$ encodes the patch-center physical time. To further capture location-dependent temporal structure, we introduce Spatial-Temporal Interactive 4D Encoding (Interactive 4D), which extends the original 4D space-time encoding with a multiplicative interaction between spatial and temporal features. This design aligns EEG and fNIRS tokens by physical time and provides the model with location-dependent temporal information, rather than relying on fixed channel indices tied to a specific montage. Second, Bio-MF adopts cross-modal classifier-free guidance (CFG) [37] to balance EEG-conditioned responses with the learned fNIRS prior. Since EEG conditions can be noisy, overly strong conditioning may transfer non-physiological EEG perturbations into fNIRS, whereas overly weak conditioning may lead to an average hemodynamic template. CFG provides a controllable mechanism for combining conditional trial information with the marginal fNIRS prior.

Experiments on paired EEG-fNIRS Dataset 1 and EEG-only Dataset 2 indicate that Bio-MF can generate task-related fNIRS signals for downstream hybrid MI decoding. Compared with SCDM, Bio-MF obtains comparable generated-modality decoding performance in most evaluated settings while reducing generation latency from seconds to milliseconds. On an RTX PRO 6000 GPU, Bio-MF generates one fNIRS trial in 7.0 ms, corresponding to an 857x speedup over the hardware-matched 1000-step SCDM baseline. These results suggest that the Bio-MF framework provides a feasible path toward fast and high-quality EEG-to-fNIRS signal generation for hybrid MI-BCIs.

The main contributions are summarized as follows:

\begin{enumerate}
\item We propose Bio-MF, a Transformer-based latent-free single-step EEG-to-fNIRS generation framework for hybrid MI-BCIs. Bio-MF directly predicts clean fNIRS signals rather than noise in the raw fNIRS signal space and uses MeanFlow velocity supervision to learn the interval-average denoising direction from a noisy fNIRS state toward the clean fNIRS signal, enabling 1-NFE inference without VAE compression or multi-step reverse diffusion.

\item We introduce Spatial-Temporal Interactive 4D Encoding and cross-modal CFG for EEG-conditioned fNIRS generation. Interactive 4D uses continuous coordinates for EEG electrodes, fNIRS channels, and physical time while modeling spatial-temporal interactions. CFG controls the balance between EEG-conditioned responses and the fNIRS prior.

\item We use a two-stage noise-level-gated FFT regularizer to address the speed-fidelity trade-off of one-step generation. By applying spectral alignment only in low-noise states, the training strategy is designed to reduce frequency-domain artifacts while keeping the millisecond-scale inference path.

\item We evaluate Bio-MF using downstream classification, cross-device zero-shot evaluation, signal-level visualizations, ablation studies, and measured generation latency. The results suggest that Bio-MF can provide task-useful synthetic fNIRS while reducing EEG-to-fNIRS generation latency compared with multi-step diffusion baselines, supporting hybrid MI-BCIs in rehabilitation training, assistive control, and feedback-based motor-imagery practice.
\end{enumerate}

\section{Datasets}

\subsection{Dataset 1}

Dataset 1 is a publicly available hybrid motor imagery dataset introduced by Shin et al. [27], containing temporally aligned EEG and fNIRS recordings. This dataset was used as the paired source domain for training the proposed EEG-to-fNIRS generator and for evaluating in-distribution downstream classification performance. It includes 29 healthy participants performing left- and right-hand motor imagery (LMI/RMI) tasks. Each participant completed three sessions, with 20 trials per session, resulting in 1740 valid MI trials. Each trial consisted of a preparation period, a task execution interval, and a subsequent resting period, allowing the recorded fNIRS signals to cover the delayed hemodynamic response induced by motor imagery.

EEG was recorded from 30 active electrodes arranged according to the international 10-5 system. The EEG signals were resampled to 160 Hz, band-pass filtered between 0.5 and 50 Hz using a fourth-order Chebyshev type-II filter, and further processed with independent component analysis (ICA) to reduce ocular artifacts. fNIRS was acquired using 14 light sources and 16 detectors, forming 36 measurement channels. The fNIRS signals were downsampled to 10 Hz, filtered with a sixth-order zero-phase Butterworth band-pass filter within 0.01-0.1 Hz, and converted into oxy-hemoglobin (HbO) and deoxy-hemoglobin (HbR) concentration changes. Before model input, EEG and fNIRS tensors were z-score normalized to reduce subject-, channel-, and session-level scale differences. For model training and evaluation, each EEG trial was represented as a $30 \times 4000$ sequence, and each fNIRS trial was represented as a two-channel tensor of size $2 \times 36 \times 256$, where the two channels correspond to HbO and HbR.

\subsection{Dataset 2}

Dataset 2 was introduced to assess whether Bio-MF can generalize to an unseen EEG acquisition setting without paired fNIRS supervision. This dataset was derived from a PhysioNet motor imagery EEG resource recorded with BCI2000 [28], [29] and contains only EEG recordings. Therefore, it was not used for generator training or fNIRS reconstruction supervision. Instead, it provides a strict zero-shot evaluation scenario in which the trained generator receives EEG from a different device layout and produces synthetic fNIRS signals for downstream hybrid MI-BCI classification.

Following the SCDM evaluation protocol [25], we selected data from 100 subjects performing LMI/RMI tasks. Each subject contributed 42 trials with a balanced LMI/RMI ratio, yielding 4200 samples. Unlike Dataset 1, which uses a 30-electrode EEG montage based on the international 10-5 system, Dataset 2 uses a 64-channel high-density EEG montage following the international 10-10 system. The EEG signals were sampled at 160 Hz, filtered within 8-30 Hz using a finite impulse response (FIR) band-pass filter, and z-score normalized before zero-shot fNIRS generation and downstream classification. For Dataset 2, Bio-MF uses the 64-channel EEG coordinates as the condition tokens and generates synthetic HbO/HbR fNIRS on the 36-channel fNIRS layout learned from Dataset 1.

\section{Methodology}

\begin{figure*}[!t]
\centering
\includegraphics[width=0.82\textwidth]{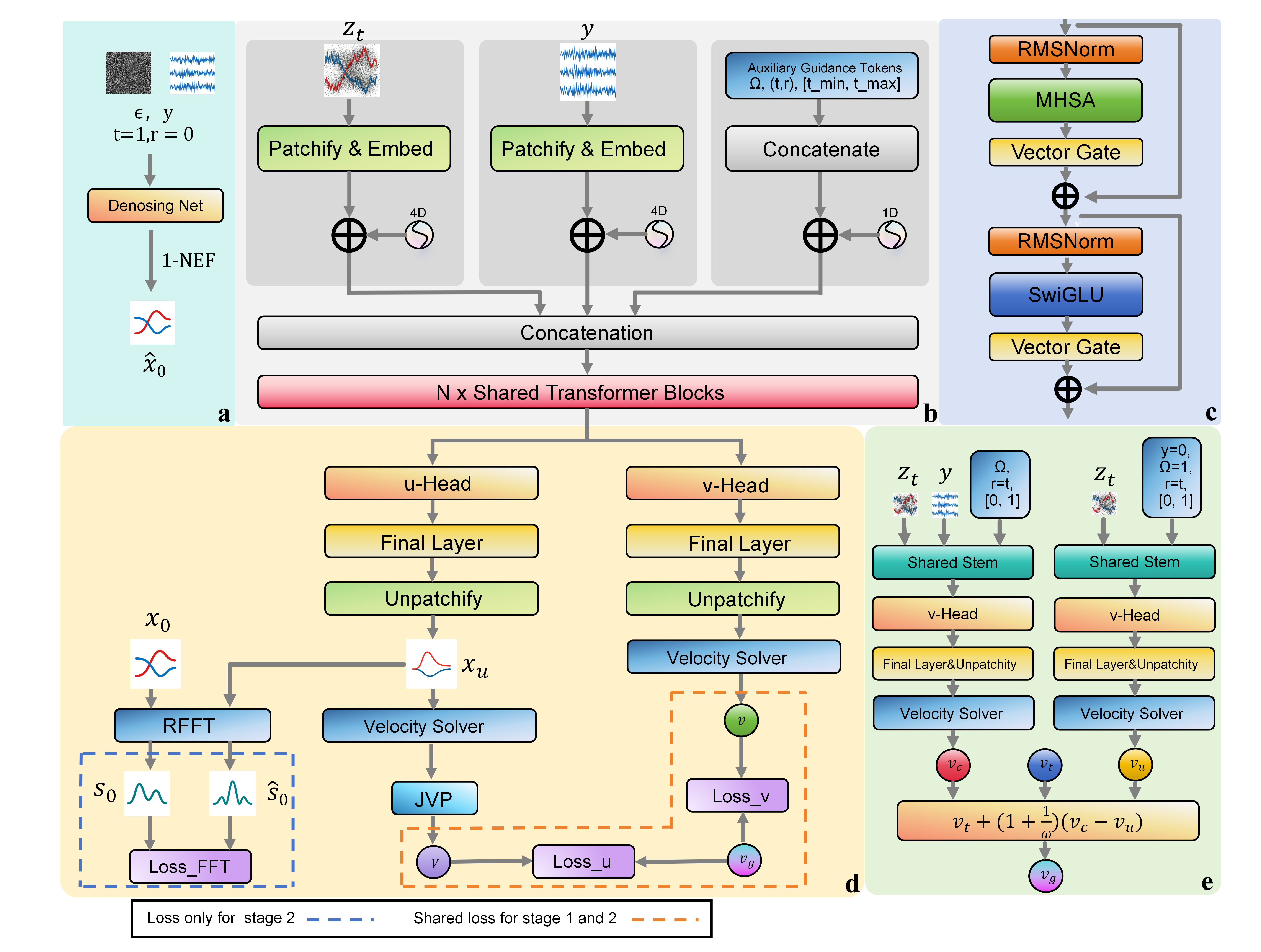}
\caption{Overall architecture of Bio-MF. (a) One-step fNIRS generation from the sampled noise endpoint and EEG condition. (b) Multimodal tokenization and shared Transformer blocks. (c) Transformer block structure. (d) Dual-branch MeanFlow velocity supervision and noise-level-gated FFT regularization. (e) CFG velocity target construction using the same shared-stem structure as in (b).}
\end{figure*}

\subsection{Overall Architecture of the Proposed Framework}

As shown in Fig. 1, Bio-MF uses a tokenized Transformer backbone shared by training and one-step inference. EEG and noised fNIRS are divided into physically aligned multi-rate patches and augmented with continuous sensor-time coordinates and auxiliary control tokens. During training, a shared stem and separate $u$- and $v$-heads are optimized with MeanFlow velocity supervision, CFG target construction, and noise-level-gated spectral regularization. The $u$-head predicts a clean-signal field that is converted to MeanFlow velocity, while the training-only $v$-head supplies auxiliary velocity supervision and CFG targets [30], [31], [37]. At inference, $t=1$ and $r=0$, and the shared stem plus $u$-head generate the final fNIRS signal with one network evaluation.

Each Transformer block contains RMSNorm, multi-head self-attention, SwiGLU, and vector-gated residual connections [32], [38], [39]. Interactive 4D encoding represents sensor geometry and physical patch time [35], [36], while the second training stage applies a gated FFT loss to constrain low-frequency fNIRS morphology. The following subsections define these components.

\subsection{Spatial-Temporal Interactive 4D Encoding and Multi-Rate Tokenization}

Conventional EEG deep learning studies often treat electrode channels as a one-dimensional array or project them onto a two-dimensional planar grid, such as a $16 \times 16$ matrix. These channel representations can distort the native anatomical geometry of the brain and disrupt true spatial distances between recording sites, making models less robust when electrode layouts change, such as from 30 to 64 channels. The REVE EEG foundation model shows that jointly encoding three-dimensional electrode coordinates and time indices as 4D positional encoding can reduce fixed-montage dependence and improve generalization to unseen recording setups [36]. Inspired by this idea, Bio-MF extends the 4D representation from EEG representation learning to EEG-to-fNIRS cross-modal generation.

To avoid such geometric distortion, Bio-MF discards discrete channel indices and instead represents EEG and fNIRS measurements with continuous spatial coordinates and physical time. Because EEG and fNIRS have substantially different sampling rates, we first apply a multi-rate tokenization strategy: EEG signals at 160 Hz and fNIRS signals at 10 Hz are divided according to matched physical time windows, with fNIRS represented by 0.8-s temporal patches. EEG tokens are partitioned using the corresponding physical intervals, and each EEG/fNIRS token is assigned its physical patch-center timestamp. The two modalities are thereby aligned by physical time rather than by raw sample indices and represented as unified sequential features.

Before entering the denoising network, EEG tokens and noisy fNIRS tokens each incorporate the proposed Interactive 4D positional encoding. The guidance scale $\omega$, guidance interval [$t_{\min}$,$t_{\max}$], and MeanFlow time pair $(t,r)$ are embedded as auxiliary control tokens. The three types of tokens are then concatenated and fed into the Transformer denoising network.

After patching, each token is assigned an explicit physical coordinate vector

\begin{equation}
\boldsymbol{c}_i=(x_i,y_i,z_i,t_i),
\end{equation}

where $(x_i,y_i,z_i)$ denotes the three-dimensional Euclidean position of the corresponding electrode or optical measurement channel in a standard head model, and $t_i$ denotes the physical center timestamp of the temporal patch in seconds.

The original 4D encoding was introduced for EEG-centric representation learning, where tokens are mainly organized within a single electrophysiological modality. It represents each token using a joint $(x,y,z,t)$ coordinate, which provides continuous spatial-temporal information but does not explicitly separate spatial and temporal factors or parameterize their multiplicative interaction. For EEG-to-fNIRS generation, however, the model must bridge two heterogeneous modalities with different sensor layouts, sampling rates, and neurophysiological dynamics. Therefore, the original joint 4D design may be suboptimal for representing cross-modal spatial-temporal coupling between EEG conditions and fNIRS responses. To address this limitation, Interactive 4D decouples spatial and temporal Fourier features and introduces an explicit interaction term, making the coordinate representation better suited to multimodal EEG-to-fNIRS generation.

Following the general design principle of REVE [36], the positional representation contains a structured Fourier branch and a lightweight learnable coordinate branch, followed by normalization after fusion. In the Fourier branch, we decouple the spatial coordinates and the temporal coordinate before introducing the explicit spatial-temporal interaction term. Let $\boldsymbol{p}_i=(x_i,y_i,z_i)$ be the spatial coordinate. We first compute separate spatial and temporal Fourier features:

\begin{equation}
\boldsymbol{s}_i=\Phi_s(\boldsymbol{p}_i),
\quad
\boldsymbol{q}_i=\Phi_t(t_i),
\end{equation}

where $\Phi_s(\cdot)$ jointly encodes the three-dimensional sensor position using spatial Fourier bases, and $\Phi_t(\cdot)$ encodes the physical patch-center time. The Fourier mapping follows the standard sinusoidal form. For a scalar coordinate projection $c$ and frequency $f_k$, one basis component is

\begin{equation}
\psi_k(c)=
\left[
\cos\left(c f_k \frac{2\pi}{W}\right),
\sin\left(c f_k \frac{2\pi}{W}\right)
\right],
\end{equation}

where $W$ is a normalization constant for the coordinate range. In the spatial branch, $c$ is obtained by projecting $\boldsymbol{p}_i$ onto a spatial Fourier direction; in the temporal branch, $c=t_i$.

To model location-dependent temporal effects, Interactive 4D introduces a multiplicative interaction between the spatial and temporal Fourier features:

\begin{equation}
\boldsymbol{r}_i=(W_s\boldsymbol{s}_i)\odot(W_t\boldsymbol{q}_i),
\end{equation}

where $W_s$ and $W_t$ are learnable linear projections and $\odot$ denotes element-wise multiplication. The Fourier-based positional branch is then defined as

\begin{equation}
\boldsymbol{F}_{pe,i}=\boldsymbol{s}_i+\boldsymbol{q}_i+\boldsymbol{r}_i.
\end{equation}

Here, $\boldsymbol{F}_{pe,i}$ is not a second Fourier transform; it is the fused output of the spatial Fourier, temporal Fourier, and interactive Fourier-based features.

Following the REVE positional-fusion design [36], we also include a learnable coordinate branch to adapt the continuous coordinates to the target generation setting:

\begin{equation}
\boldsymbol{F}_{lin,i}=\mathrm{LayerNorm}\left(\mathrm{GELU}\left(W_l\boldsymbol{c}_i+b_l\right)\right).
\end{equation}

The final 4D positional representation is obtained by fusing the Fourier-based branch and the learnable branch:

\begin{equation}
\boldsymbol{p}^{4D}_i=
\mathrm{LayerNorm}\left(\boldsymbol{F}_{pe,i}+\boldsymbol{F}_{lin,i}\right).
\end{equation}

Finally, the positional representation is added to the corresponding signal token:

\begin{equation}
\boldsymbol{h}_i=\boldsymbol{h}_i^{signal}+\boldsymbol{p}^{4D}_i.
\end{equation}

\subsection{Latent-Free MeanFlow for 1-Step Generation}

Bio-MF does not use a VAE encoder/decoder and does not primarily output a noise residual. Instead, it directly predicts the clean fNIRS signal in the raw fNIRS signal space, and then converts this signal-form output into velocity space for supervision. This design avoids reconstruction bias introduced by latent compression and keeps the model output interpretable as an fNIRS signal.

Unlike $\epsilon$-prediction or score-prediction, direct clean-signal prediction parameterizes the target data itself, so the network output remains in an interpretable physiological signal space. This design allows structural and task-related constraints to be imposed directly on the generated content, including channel consistency, temporal smoothness, frequency-domain consistency, or prior mask constraints, without indirectly mapping these constraints into noise or score space. For EEG-to-fNIRS generation, where the synthetic signal should preserve clear physiological and task semantics, direct signal prediction therefore provides a more natural way to jointly optimize generation quality and one-step inference speed.

During training, given a real fNIRS signal $\mathbf{x}_0$ and Gaussian noise $\boldsymbol{\epsilon}\sim\mathcal{N}(\mathbf{0},\mathbf{I})$, the forward interpolation state is defined as

\begin{equation}
\label{eq:interpolation_state}
\mathbf{z}_t=(1-t)\mathbf{x}_0+t\boldsymbol{\epsilon}, \qquad t\in[0,1].
\end{equation}

Here, $\mathbf{x}_0\sim p_{\mathrm{data}}$ and $\boldsymbol{\epsilon}\sim p_{\mathrm{prior}}$; thus $\mathbf{z}_0=\mathbf{x}_0\sim p_{\mathrm{data}}$ and $\mathbf{z}_1=\boldsymbol{\epsilon}\sim p_{\mathrm{prior}}$. MeanFlow does not integrate along the full ODE trajectory step by step. Instead, it learns the interval average velocity $\mathbf{u}$ from the current time $t$ to the target time $r$. The velocity expressions below are evaluated for $t>0$ in practice, with $t$ sampled away from zero to avoid the degenerate endpoint.

The Bio-MF $u$-head first outputs the signal-form field $\mathbf{x}_u$, where $\mathbf{x}_u$ denotes the network-generated generalized clean-signal field induced by the $u$-head under the current $(\mathbf{z}_t,t,r)$ condition. When $r=0$, this generalized output corresponds to the clean endpoint signal $\mathbf{x}_0$. This signal-form output is then converted into average velocity:

\begin{equation}
\label{eq:meanflow_velocity}
\mathbf{u}=\frac{\mathbf{z}_t-\mathbf{x}_u}{t}.
\end{equation}

Because $\mathbf{u}$ represents the network-output interval average velocity while the training target is defined in velocity space, Bio-MF uses the MeanFlow identity to construct a supervised composite velocity [31]:

\begin{equation}
\label{eq:meanflow_identity}
\mathbf{V}=\mathbf{u}+(t-r)\cdot\mathbf{JVP}_{\mathrm{sg}}.
\end{equation}

where JVP denotes the Jacobian-vector product used to compute $d\mathbf{u}/dt$, and ``sg'' denotes stop-gradient; $\mathbf{V}$ is the MeanFlow consistency velocity used to align the velocity target during training, rather than a new network output. The guided velocity target $\mathbf{v}_g$ is defined by the cross-modal guidance mechanism in the next subsection.

\subsection{Cross-Modal CFG}

Bio-MF uses cross-modal CFG as a statistical conditioning mechanism rather than a causal model of neurovascular coupling. During training, the auxiliary $v$-head evaluates real and zero EEG under the zero-interval setting $r=t$ to obtain conditional and unconditional velocities:

\begin{equation}
\label{eq:conditional_unconditional_velocity}
\resizebox{0.84\columnwidth}{!}{$\displaystyle
\mathbf{v}_c=f_\theta^v(\mathbf{z}_t,t,t,\omega,0,1,\mathbf{y}),
\quad
\mathbf{v}_u=f_\theta^v(\mathbf{z}_t,t,t,1,0,1,\mathbf{0}).$}
\end{equation}

Here, $\mathbf{v}_c$ is the EEG-conditioned direction and $\mathbf{v}_u$ is the marginal fNIRS-prior direction. Given the base velocity

\begin{equation}
\label{eq:base_flow_velocity}
\mathbf{v}_t=\frac{\mathbf{z}_t-\mathbf{x}_0}{t},
\end{equation}

the guided velocity target is defined as

\begin{equation}
\label{eq:guided_velocity_target}
\mathbf{v}_g=\mathbf{v}_t+\left(1-\frac{1}{\omega_{int}}\right)(\mathbf{v}_c-\mathbf{v}_u).
\end{equation}

where $\omega_{int}=\omega$ for $t\in[t_{\min},t_{\max}]$ and $\omega_{int}=1$ otherwise. The offset $\mathbf{v}_c-\mathbf{v}_u$ captures EEG conditioning relative to the learned fNIRS prior, not a causal hemoglobin increment. The $u$-head learns the JVP-corrected average velocity $\mathbf{V}$, whereas the $v$-head directly regresses the same target. Their per-sample errors are

\begin{equation}
\label{eq:velocity_losses}
\ell_u=\left\|\mathbf{V}-\mathbf{v}_g\right\|_2^2, \qquad
\ell_v=\left\|\mathbf{v}-\mathbf{v}_g\right\|_2^2.
\end{equation}

The MeanFlow loss combines both velocity errors:

\begin{equation}
\label{eq:meanflow_loss}
\mathcal{L}_{MF}=\mathbb{E}\left[\ell_u+\ell_v\right].
\end{equation}

The $v$-head and CFG target construction are omitted at inference.

\subsection{Two-Stage Optimization with Noise-Level-Gated Spectral Regularization}

To suppress high-frequency artifacts without disturbing high-noise velocity learning, Stage I optimizes only $\mathcal{L}_{MF}$, whereas Stage II forms $\mathbf{x}_u=\mathbf{z}_t-t\mathbf{u}$ and applies FFT regularization only for $t<0.4$:

\begin{equation}
\label{eq:fft_loss}
\mathcal{L}_{fft} = \left\| \left|\mathcal{F}_{t}(\mathbf{x}_u)\right| - \left|\mathcal{F}_{t}(\mathbf{x}_0)\right| \right\|_2^2.
\end{equation}

Here, $\mathcal{F}_{t}$ is a one-dimensional FFT along time, and the norm spans HbO/HbR, spatial channels, and frequencies. The Stage II objective is

\begin{equation}
\label{eq:total_loss}
\mathcal{L}_{total} = \mathbb{E}\left[\ell_u+\ell_v+\lambda_{fft} \cdot \mathbb{I}(t<0.4)\mathcal{L}_{fft}\right].
\end{equation}

The low-noise gate limits conflicts between spectral alignment and high-noise velocity learning. Fig. 2 contrasts the nonsmooth outputs obtained without FFT loss with the periodic artifacts produced by globally applying the spectral term.

\begin{figure*}[!t]
\centering
\includegraphics[width=0.82\textwidth]{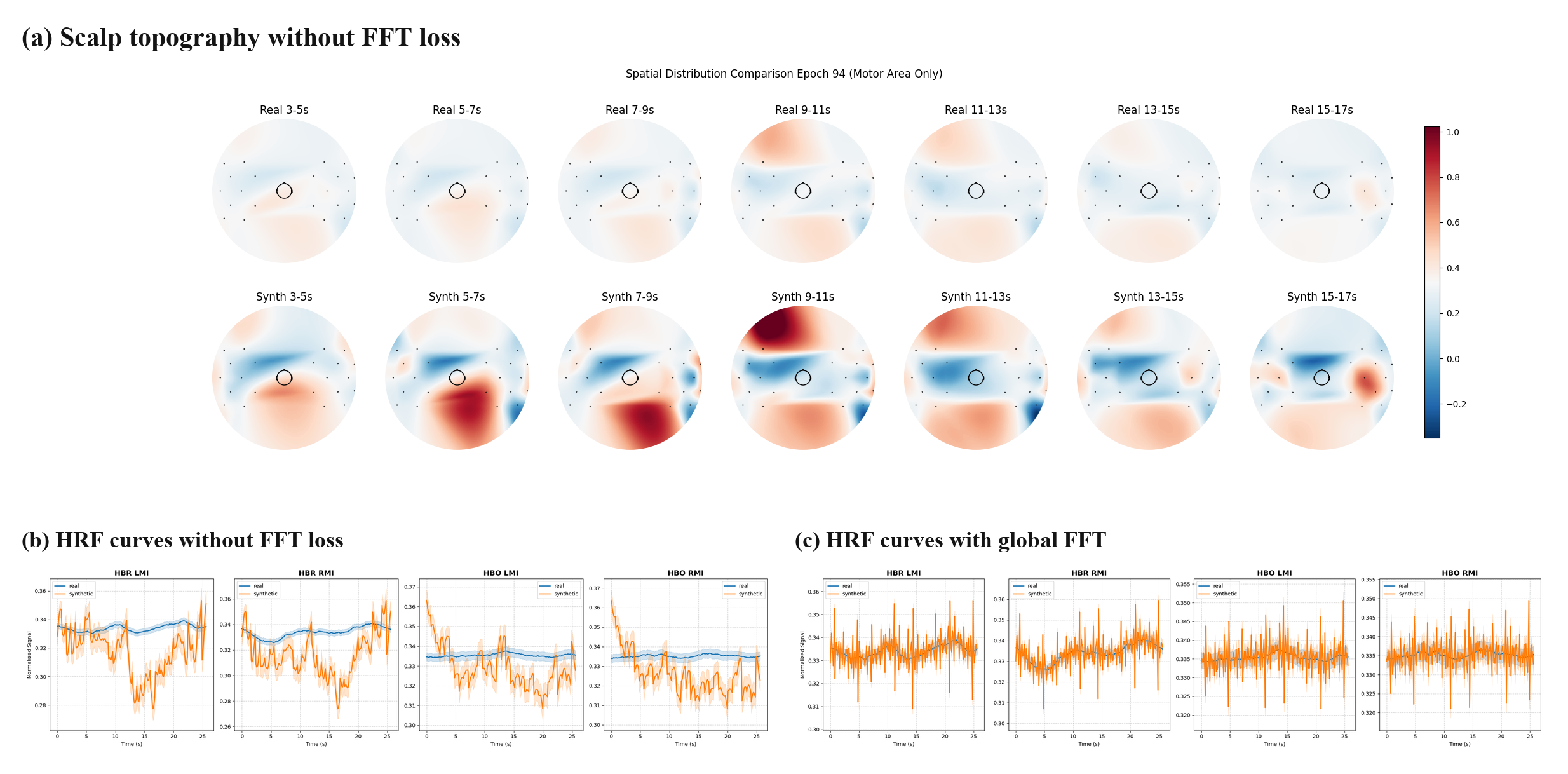}
\caption{Diagnostic visualizations motivating two-stage noise-level-gated FFT regularization. (a) Scalp topographies without FFT loss show locally nonsmooth synthetic fNIRS responses compared with real fNIRS. (b) HRF curves without FFT loss show high-frequency jitter and morphology deviation in synthetic fNIRS. (c) HRF curves under global FFT regularization, without the two-stage schedule and noise-level gate, show periodic or impulse-like artifacts.}
\end{figure*}

\begin{algorithm}[htbp]
\caption{Training Phase of Bio-MF}
\label{alg:bio_mf_training}
\setlength{\fboxsep}{3pt}
\setlength{\fboxrule}{0.4pt}
\noindent\fbox{%
\begin{minipage}{\dimexpr\linewidth-2\fboxsep-2\fboxrule\relax}
\scriptsize
\begin{algorithmic}
\STATE \textbf{Define:}
\STATE \quad $\mathbf{x}_0$: real fNIRS; $\mathbf{y}$: EEG condition; $\boldsymbol{\epsilon}$: Gaussian noise
\STATE \quad $\mathcal{F}_{\theta}=(\mathcal{F}_{\theta,u},\mathcal{F}_{\theta,v})$: Bio-MF denoising network
\STATE \textbf{Input:}
\STATE \quad $\mathbf{x}_0,\mathbf{y}$
\STATE \textbf{State Construction:}
\STATE \quad Sample $t>0,r,\omega,t_{\min},t_{\max}$ and $\boldsymbol{\epsilon}\sim\mathcal{N}(\mathbf{0},\mathbf{I})$
\STATE \quad $\mathbf{z}_t\leftarrow(1-t)\mathbf{x}_0+t\boldsymbol{\epsilon}$, $\mathbf{v}_t\leftarrow(\mathbf{z}_t-\mathbf{x}_0)/t$
\STATE \textbf{MeanFlow Denoising:}
\STATE \quad $\mathbf{x}_u\leftarrow\mathcal{F}_{\theta,u}^{x}(\mathbf{z}_t,\mathbf{y},t,r,\omega,t_{\min},t_{\max})$
\STATE \quad $\mathbf{u}\leftarrow(\mathbf{z}_t-\mathbf{x}_u)/t$, $\mathbf{v}\leftarrow\mathcal{F}_{\theta,v}(\mathbf{z}_t,\mathbf{y},t,r,\omega,t_{\min},t_{\max})$
\STATE \textbf{Guided Target:}
\STATE \quad $\omega_{int}\leftarrow\omega$ if $t\in[t_{\min},t_{\max}]$, otherwise $1$
\STATE \quad $(\mathbf{v}_c,\mathbf{v}_u)\leftarrow\mathcal{F}_{\theta,v}(\mathbf{z}_t,\mathbf{y},t,t,\omega_{int},0,1),\mathcal{F}_{\theta,v}(\mathbf{z}_t,\mathbf{0},t,t,1,0,1)$
\STATE \quad $\mathbf{v}_g\leftarrow\mathbf{v}_t+(1-1/\omega_{int})(\mathbf{v}_c-\mathbf{v}_u)$
\STATE \quad If condition dropout is applied, $(\mathbf{y},\mathbf{v}_g)\leftarrow(\mathbf{0},\mathbf{v}_t)$
\STATE \textbf{Loss/Update:}
\STATE \quad Let $\mathcal{U}_{\theta,u}$ be the velocity field induced by $\mathcal{F}_{\theta,u}^{x}$ through $\mathbf{u}=(\mathbf{z}_t-\mathbf{x}_u)/t$
\STATE \quad $\mathbf{V}\leftarrow\mathbf{u}+(t-r)\operatorname{sg}[\operatorname{JVP}(\mathcal{U}_{\theta,u};(\mathbf{z}_t,t,r),(\mathbf{v}_c,1,0))]$
\STATE \quad $\mathcal{L}_{mf}\leftarrow\|\mathbf{V}-\mathbf{v}_g\|_2^2+\|\mathbf{v}-\mathbf{v}_g\|_2^2$
\STATE \quad $\mathcal{L}_{fft}\leftarrow0$; if Stage II and $t<0.4$, set $\mathcal{L}_{fft}\leftarrow\left\||\operatorname{RFFT}(\mathbf{x}_u)|-|\operatorname{RFFT}(\mathbf{x}_0)|\right\|_2^2$
\STATE \quad $\mathcal{L}_{total}\leftarrow\mathbb{E}[\mathcal{L}_{mf}+\lambda_{fft}\mathcal{L}_{fft}]$
\STATE \quad Update $\theta$ by minimizing $\mathcal{L}_{total}$
\end{algorithmic}
\end{minipage}}
\end{algorithm}

\begin{algorithm}[htbp]
\caption{Inference Phase of Bio-MF}
\label{alg:bio_mf_inference}
\setlength{\fboxsep}{3pt}
\setlength{\fboxrule}{0.4pt}
\noindent\fbox{%
\begin{minipage}{\dimexpr\linewidth-2\fboxsep-2\fboxrule\relax}
\scriptsize
\begin{algorithmic}
\STATE \textbf{Define:}
\STATE \quad $\mathcal{D}_{\theta}^{u}$: trained Bio-MF denoising network
\STATE \textbf{Input:}
\STATE \quad EEG $\mathbf{y}$, CFG scale $\omega$, interval $[t_{\min},t_{\max}]$
\STATE \textbf{Initialize Generation:}
\STATE \quad $t\leftarrow1$, $r\leftarrow0$
\STATE \textbf{Sampling Process:}
\STATE \quad Sample $\boldsymbol{\epsilon}\sim\mathcal{N}(\mathbf{0},\mathbf{I})$
\STATE \textbf{Denoising Process:}
\STATE \quad $\hat{\mathbf{x}}_0\leftarrow\mathcal{D}_{\theta}^{u}(\boldsymbol{\epsilon},\mathbf{y},t,r,\omega,t_{\min},t_{\max})$
\STATE \textbf{Output:}
\STATE \quad Reconstruct final fNIRS signal $\hat{\mathbf{x}}_0$
\end{algorithmic}
\end{minipage}}
\end{algorithm}

Bio-MF was implemented as a Transformer denoising network with hidden dimension 384 and 8 Transformer blocks. The architecture contains a 4-block shared stem followed by separate 4-block $u$- and $v$-heads, with the generation branch used at inference and the $v$-head used only during training for velocity supervision. Each block uses 6-head self-attention with 64 channels per head, RMSNorm, SwiGLU feed-forward layers with an expansion ratio of $8/3$, vector-valued residual gating, and spatiotemporal Fourier positional encoding that jointly encodes $(x,y,z)$ sensor coordinates and separately encodes the temporal coordinate $t$.

Using this configuration, all Bio-MF experiments were trained on a single NVIDIA RTX PRO 6000 GPU. The model was optimized for 240 epochs, with a 10-epoch warmup period at the beginning of training. Following the two-stage objective described above, Stage I occupied the first 150 epochs and optimized only $\mathcal{L}_{MF}$, while Stage II used the remaining 90 epochs and enabled the noise-level-gated FFT regularizer. The complete training run took 8 h. By comparison, SCDM is reported to use a 30,000-epoch training schedule [25], indicating that Bio-MF uses a substantially shorter optimization schedule in addition to its one-step inference advantage.

\section{Results}

\subsection{Classification}

To evaluate whether generated fNIRS provides useful complementary information for MI-BCI decoding, we performed downstream left/right motor imagery classification under single-modality and hybrid-modality settings. For Dataset 1, paired EEG-fNIRS trials provide the source-domain supervision for EEG-conditioned fNIRS synthesis. For Dataset 2, which contains only EEG recordings with an unseen 64-channel montage, Bio-MF was applied in a zero-shot manner to synthesize 36-channel fNIRS signals on the source fNIRS layout. Table 1 compares Bio-MF with the SCDM classification results reported under the same classifier family and ratio-based evaluation protocol [25].

To keep the comparison aligned with prior EEG-to-fNIRS generation studies, the same classifier family was used for each modality setting: ESNet for EEG-only input, FSNet for fNIRS-only input, and FGANet for EEG-fNIRS hybrid input [15], [25]. Following the SCDM evaluation protocol [25], classifiers were evaluated under seven LMI/RMI training ratios, from 1:4 to 4:1, with the remaining samples used for testing. For Dataset 1, these ratios correspond to 200/800, 300/700, 400/600, 500/500, 600/400, 700/300, and 800/200 LMI/RMI training samples drawn from 870 trials per class. The balanced 1:1 setting used 10-fold cross-validation repeated 10 times, whereas the other ratios used 20 random train/test repetitions; Dataset 2 followed the same ratio-based protocol. We report the mean and standard deviation over all runs using ACC, SPE, PRE, and SEN, where SEN and SPE measure LMI and RMI recall, respectively, and PRE measures LMI precision.

The results are summarized in Table 1. For each dataset, the Bio-MF EEG-only baseline is retained as the reference for interpreting synthetic fNIRS-only and EEG + synthetic fNIRS performance. The Bio-MF fusion rows correspond to the Full Bio-MF configuration with Interactive 4D encoding.

\begin{table*}[!t]
\caption{Bio-MF and SCDM classification results.}
\centering
\vspace{2pt}
\scriptsize
\begingroup
\setlength{\tabcolsep}{1.7pt}
\renewcommand{\arraystretch}{0.96}
\setlength{\extrarowheight}{0.7pt}
\setlength{\arrayrulewidth}{0.15pt}
\resizebox{0.96\textwidth}{!}{%
\begin{tabular}{c|cccccc}
\hline
Dataset & Signal & Model & ACC (\%) & SPE (\%) & PRE (\%) & SEN (\%) \\
\hline
\multirow{5}{*}{\textbf{Dataset 1}} & EEG & - & 71.92 $\pm$ 2.79 & 76.85 $\pm$ 3.10 & 74.36 $\pm$ 3.12 & 66.98 $\pm$ 3.48 \\
\hhline{~|------}
 & \begin{tabular}[c]{@{}c@{}}Syn. fNIRS\\(HbR)\end{tabular} & \begin{tabular}[c]{@{}c@{}}Bio-MF\\SCDM~[25]\end{tabular} & \begin{tabular}[c]{@{}c@{}}66.42 $\pm$ 2.70\\66.38 $\pm$ 2.56\end{tabular} & \begin{tabular}[c]{@{}c@{}}69.25 $\pm$ 2.86\\68.27 $\pm$ 2.48\end{tabular} & \begin{tabular}[c]{@{}c@{}}68.52 $\pm$ 3.00\\68.32 $\pm$ 2.73\end{tabular} & \begin{tabular}[c]{@{}c@{}}69.59 $\pm$ 2.78\\68.26 $\pm$ 2.86\end{tabular} \\
\hhline{~|------}
 & \begin{tabular}[c]{@{}c@{}}EEG + Syn.\\fNIRS (HbR)\end{tabular} & \begin{tabular}[c]{@{}c@{}}Bio-MF\\SCDM~[25]\end{tabular} & \begin{tabular}[c]{@{}c@{}}75.29 $\pm$ 2.36\\75.38 $\pm$ 2.23\end{tabular} & \begin{tabular}[c]{@{}c@{}}80.09 $\pm$ 2.61\\79.98 $\pm$ 2.33\end{tabular} & \begin{tabular}[c]{@{}c@{}}77.65 $\pm$ 2.72\\77.52 $\pm$ 2.61\end{tabular} & \begin{tabular}[c]{@{}c@{}}70.22 $\pm$ 2.79\\70.50 $\pm$ 2.71\end{tabular} \\
\hhline{~|------}
 & \begin{tabular}[c]{@{}c@{}}Syn. fNIRS\\(HbO)\end{tabular} & \begin{tabular}[c]{@{}c@{}}Bio-MF\\SCDM~[25]\end{tabular} & \begin{tabular}[c]{@{}c@{}}70.12 $\pm$ 2.82\\68.52 $\pm$ 2.97\end{tabular} & \begin{tabular}[c]{@{}c@{}}77.70 $\pm$ 2.70\\76.33 $\pm$ 2.71\end{tabular} & \begin{tabular}[c]{@{}c@{}}73.05 $\pm$ 3.12\\71.79 $\pm$ 3.24\end{tabular} & \begin{tabular}[c]{@{}c@{}}60.40 $\pm$ 2.76\\60.25 $\pm$ 2.74\end{tabular} \\
\hhline{~|------}
 & \begin{tabular}[c]{@{}c@{}}EEG + Syn.\\fNIRS (HbO)\end{tabular} & \begin{tabular}[c]{@{}c@{}}Bio-MF\\SCDM~[25]\end{tabular} & \begin{tabular}[c]{@{}c@{}}76.07 $\pm$ 2.40\\75.03 $\pm$ 2.55\end{tabular} & \begin{tabular}[c]{@{}c@{}}82.59 $\pm$ 2.61\\81.92 $\pm$ 3.07\end{tabular} & \begin{tabular}[c]{@{}c@{}}80.01 $\pm$ 2.76\\79.32 $\pm$ 3.41\end{tabular} & \begin{tabular}[c]{@{}c@{}}69.50 $\pm$ 2.85\\67.32 $\pm$ 2.31\end{tabular} \\
\hline
\multirow{5}{*}{\textbf{Dataset 2}} & EEG & - & 70.78 $\pm$ 5.79 & 71.83 $\pm$ 5.69 & 71.56 $\pm$ 5.73 & 69.32 $\pm$ 6.77 \\
\hhline{~|------}
 & \begin{tabular}[c]{@{}c@{}}Syn. fNIRS\\(HbR)\end{tabular} & \begin{tabular}[c]{@{}c@{}}Bio-MF\\SCDM~[25]\end{tabular} & \begin{tabular}[c]{@{}c@{}}71.83 $\pm$ 6.22\\71.17 $\pm$ 9.18\end{tabular} & \begin{tabular}[c]{@{}c@{}}73.80 $\pm$ 6.12\\73.32 $\pm$ 11.78\end{tabular} & \begin{tabular}[c]{@{}c@{}}73.55 $\pm$ 6.37\\72.31 $\pm$ 10.16\end{tabular} & \begin{tabular}[c]{@{}c@{}}70.90 $\pm$ 6.55\\69.77 $\pm$ 8.34\end{tabular} \\
\hhline{~|------}
 & \begin{tabular}[c]{@{}c@{}}EEG + Syn.\\fNIRS (HbR)\end{tabular} & \begin{tabular}[c]{@{}c@{}}Bio-MF\\SCDM~[25]\end{tabular} & \begin{tabular}[c]{@{}c@{}}73.76 $\pm$ 5.45\\72.18 $\pm$ 9.79\end{tabular} & \begin{tabular}[c]{@{}c@{}}74.51 $\pm$ 5.62\\73.56 $\pm$ 10.18\end{tabular} & \begin{tabular}[c]{@{}c@{}}74.21 $\pm$ 5.56\\73.32 $\pm$ 10.93\end{tabular} & \begin{tabular}[c]{@{}c@{}}71.77 $\pm$ 5.60\\70.45 $\pm$ 9.32\end{tabular} \\
\hhline{~|------}
 & \begin{tabular}[c]{@{}c@{}}Syn. fNIRS\\(HbO)\end{tabular} & \begin{tabular}[c]{@{}c@{}}Bio-MF\\SCDM~[25]\end{tabular} & \begin{tabular}[c]{@{}c@{}}71.35 $\pm$ 6.80\\70.81 $\pm$ 12.11\end{tabular} & \begin{tabular}[c]{@{}c@{}}73.35 $\pm$ 6.21\\71.61 $\pm$ 17.02\end{tabular} & \begin{tabular}[c]{@{}c@{}}73.10 $\pm$ 6.85\\72.56 $\pm$ 14.71\end{tabular} & \begin{tabular}[c]{@{}c@{}}70.75 $\pm$ 6.07\\70.01 $\pm$ 9.02\end{tabular} \\
\hhline{~|------}
 & \begin{tabular}[c]{@{}c@{}}EEG + Syn.\\fNIRS (HbO)\end{tabular} & \begin{tabular}[c]{@{}c@{}}Bio-MF\\SCDM~[25]\end{tabular} & \begin{tabular}[c]{@{}c@{}}73.28 $\pm$ 5.25\\71.31 $\pm$ 10.56\end{tabular} & \begin{tabular}[c]{@{}c@{}}74.73 $\pm$ 5.40\\73.56 $\pm$ 12.59\end{tabular} & \begin{tabular}[c]{@{}c@{}}74.15 $\pm$ 5.72\\72.81 $\pm$ 11.98\end{tabular} & \begin{tabular}[c]{@{}c@{}}71.39 $\pm$ 5.84\\69.35 $\pm$ 9.73\end{tabular} \\
\hline
\end{tabular}%
}
\endgroup
\vspace{1pt}
\end{table*}

Across the generated-modality ACC comparisons in Table 1, Bio-MF is ahead of SCDM in nearly all Dataset 1 settings. The only ACC exception is EEG + synthetic fNIRS (HbR), where Bio-MF is lower by 0.09 percentage points (75.29\% vs. 75.38\%); in HbO fusion, Bio-MF remains higher by 1.04 percentage points (76.07\% vs. 75.03\%). Averaged over HbR/HbO fusion, Bio-MF still leads SCDM by 0.47 percentage points (75.68\% vs. 75.21\%). Within the HbR fusion row, SCDM is higher only in ACC and SEN, whereas Bio-MF remains higher in SPE and PRE and leads all fNIRS-only values and all HbO-fusion values.

Dataset 2 provides a more challenging zero-shot evaluation because it contains only EEG recordings, uses a different 64-channel electrode montage, and has no paired fNIRS measurements for adaptation. In this cross-device setting, Bio-MF outperforms SCDM in every generated-modality ACC, SPE, PRE, and SEN comparison. For fusion decoding, Bio-MF exceeds SCDM by 1.58 and 1.97 percentage points in HbR and HbO ACC, with smaller ACC standard deviations (5.45/5.25 vs. 9.79/10.56). The Bio-MF fusion gains over EEG-only reach 2.98 and 2.50 percentage points, compared with 1.40 and 0.53 for SCDM.

\begin{table*}[!t]
\caption{Generation latency comparison.}
\centering
\small
\resizebox{\textwidth}{!}{%
\begin{tabular}{llllll}
\toprule
Model & Sampling state & Generated sample & Steps / NFE & Latency & Speedup \\
\midrule
SCDM [25] & Raw fNIRS signal & 2 $\times$ 36 $\times$ 256 & 1000 & 6.0 s (measured, RTX PRO 6000) & 1x \\
TADM [26] & 512-D latent diffusion & $C_{fNIRS}$ $\times$ 1000 sequence; 36 $\times$ 1000 for 36-channel datasets & 1000 & 0.5 s (reported, NVIDIA A800) & $12\times^{\dagger}$ \\
Bio-MF & Raw fNIRS signal, latent-free & 2 $\times$ 36 $\times$ 256 & 1 & 7.0 ms (measured, RTX PRO 6000) & 857x \\
\bottomrule
\end{tabular}%
}
\vspace{2pt}
\parbox{\textwidth}{\scriptsize\textit{$^{\dagger}$TADM latency is the paper-reported value for one fNIRS sequence of length 1000 under the reported NVIDIA A800 experimental setting and is not hardware-matched to our RTX PRO 6000 profiling; $C_{fNIRS}$ denotes the dataset-specific fNIRS channel count.}}
\end{table*}

\subsection{Generation Latency}

To assess generation latency relevant to hybrid MI-BCI systems, we conducted an inference profiling experiment [13], [14]. For the hardware-matched comparison, Bio-MF and SCDM were evaluated under the same input setting, where one EEG trial was used to generate one fNIRS sequence with shape $2 \times 36 \times 256$. For SCDM, the diffusion noising-step search described above selected a Wasserstein-minimum schedule with $T=1000$ and a linear beta range from $1\times10^{-5}$ to $0.015$ [25]. During inference, SCDM requires 1000 serial denoising evaluations for each generated fNIRS trial. In contrast, Bio-MF follows the single-step MeanFlow generation procedure and synthesizes each trial with one network function evaluation (1-NFE) [31]. We additionally include TADM [26] as an external reported reference: TADM performs diffusion in a 512-dimensional latent representation and reports 0.5 s for generating one fNIRS sequence of length 1000, with the final fNIRS channel count depending on the dataset-specific sensor layout.

We profiled the implemented Bio-MF and SCDM inference pipelines on the same NVIDIA RTX PRO 6000 GPU, and report the average per-trial generation time from repeated runs under the same input configuration. As summarized in Table 2, the latency gap mainly comes from the number of serial denoising evaluations: SCDM repeats its denoising pass 1000 times, whereas Bio-MF generates each trial with one network function evaluation.

The profiling results indicate that latency is dominated by the number and location of serial denoising evaluations rather than by model size alone. SCDM performs a long reverse chain directly in the raw fNIRS signal space, causing the 1000-step schedule to accumulate to 6.0 s per trial on the RTX PRO 6000. TADM reduces the per-step state to a compact 512-dimensional latent representation, which explains why its reported 1000-step latency is much lower despite still using iterative diffusion. Bio-MF avoids both the latent reconstruction path and the long reverse chain: it generates the complete raw fNIRS trial with one network function evaluation, reducing measured total generation time to 7.0 ms per trial on the RTX PRO 6000. This corresponds to an 857x generation-speed advantage over the hardware-matched SCDM baseline and provides a practical latency margin for hybrid MI-BCI feedback.

\subsection{Physiological Visualization}

We further examine whether Bio-MF preserves plausible fNIRS morphology from spatial, temporal, and topographic perspectives. For spatial correspondence, EEG signals are resampled to the fNIRS temporal length, and each fNIRS channel is connected to the EEG electrode with the highest mean absolute Pearson correlation over Dataset 1. The spatial visualization shows that the synthetic fNIRS correspondences broadly follow those of real fNIRS, indicating that generation does not collapse to arbitrary channel patterns.

\begin{figure}[!t]
\centering
\includegraphics[width=\columnwidth]{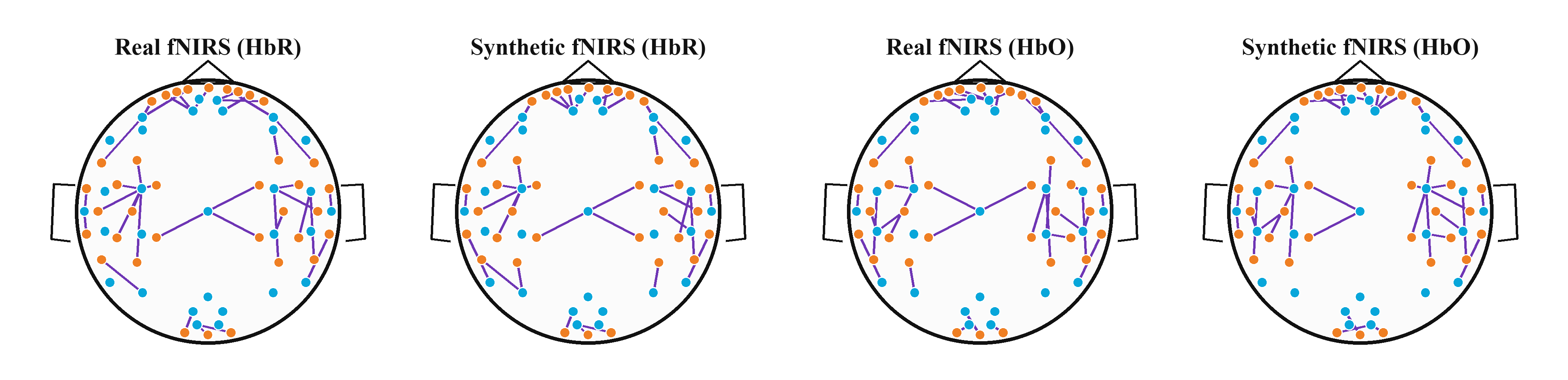}
\caption{EEG-fNIRS spatial correspondence visualization on Dataset 1. Blue markers denote EEG electrodes, orange markers denote fNIRS recording sites, and purple lines connect each fNIRS site to the EEG electrode with the highest average absolute Pearson correlation.}
\end{figure}

For temporal plausibility, trial-averaged HRF curves are computed for real and synthetic fNIRS under LMI/RMI conditions. The HRF visualization shows that synthetic curves follow the slow low-frequency trends and cross-trial variation ranges of real fNIRS in the main hemodynamic response interval, without obvious high-frequency oscillation or mean-curve collapse.

\begin{figure*}[!t]
\centering
\includegraphics[width=0.80\textwidth]{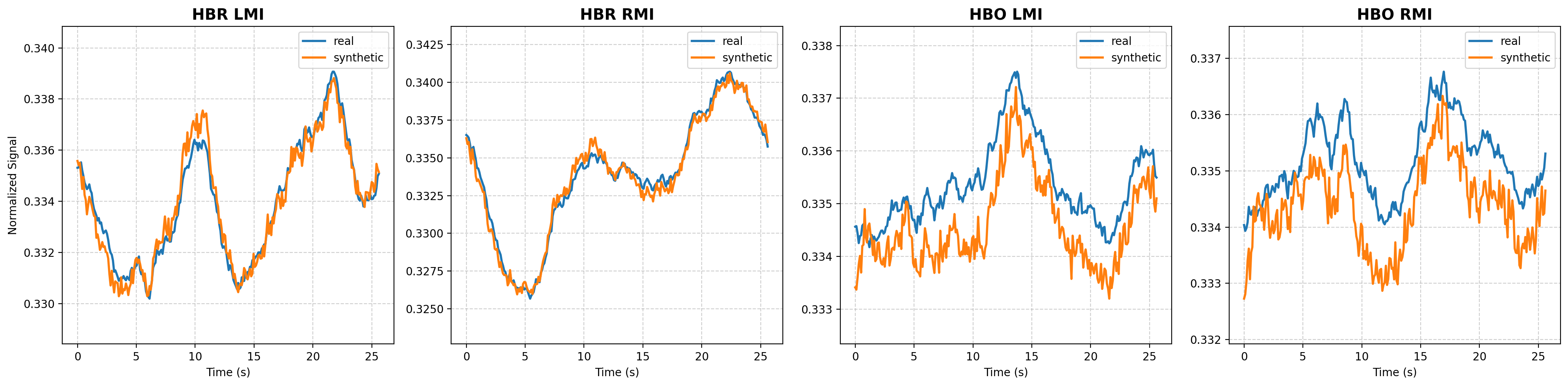}
\caption{HRF-curve visualization on Dataset 1. The panels compare average hemodynamic response curves and cross-trial variation ranges of real and synthetic fNIRS under LMI/RMI conditions.}
\end{figure*}

For scalp topology, the topography visualization compares real and synthetic fNIRS across consecutive hemodynamic windows. Synthetic fNIRS preserves the major activation regions, spatial gradients, and temporal evolution trends of real fNIRS, although local amplitude differences remain. Together with Table 1, these visualizations support the use of Bio-MF-generated fNIRS as an auxiliary modality for hybrid MI-BCI decoding.

\begin{figure*}[!t]
\centering
\includegraphics[width=0.78\textwidth]{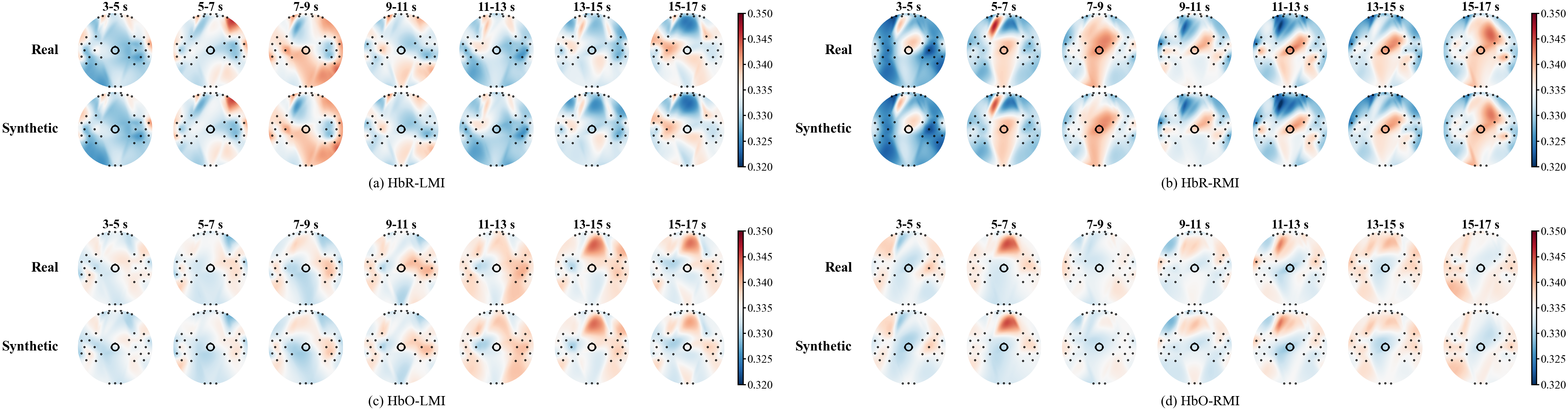}
\caption{Scalp-topography visualization on Dataset 1. This visualization compares real and synthetic fNIRS topographies for HbR/HbO under LMI/RMI conditions, evaluating whether synthetic fNIRS preserves the main activation regions, spatial gradients, and temporal evolution trends of real fNIRS.}
\end{figure*}

Additional analyses of coordinate sensitivity and CFG strength are provided in the supplementary appendix.

\section{Ablation Study}

To evaluate the contribution of each core Bio-MF module to final performance, we compare Full Bio-MF with variants that modify one design choice at a time: Original 4D replaces Interactive 4D with the original 4D encoding; w/o 4D PE removes physical coordinate encoding; w/o CFG disables the conditional guidance offset; w/o FFT removes spectral regularization; and Global FFT applies FFT regularization at all noise levels. Table 3 reports downstream classification results under the EEG + Synthetic fNIRS setting, averaged over HbR/HbO. Detailed variant definitions are provided in the supplementary appendix.

\begin{table}[H]
\caption{Classification ablation results (\%, EEG + Synthetic fNIRS; HbR/HbO average).}
\centering
\scriptsize
\setlength{\tabcolsep}{1.2pt}
\renewcommand{\arraystretch}{0.92}
\resizebox{\columnwidth}{!}{%
\begin{tabular}{lcccc}
\toprule
Variant & ACC & SPE & PRE & SEN \\
\midrule
\multicolumn{5}{l}{\textbf{Dataset 1}} \\
Full Bio-MF & \textbf{75.68 $\pm$ 2.38} & \textbf{81.34 $\pm$ 2.61} & \textbf{78.83 $\pm$ 2.74} & \textbf{69.86 $\pm$ 2.82} \\
Original 4D & 74.92 $\pm$ 2.55 & 80.62 $\pm$ 2.82 & 78.05 $\pm$ 2.95 & 69.08 $\pm$ 3.03 \\
w/o 4D PE & 74.66 $\pm$ 3.65 & 78.61 $\pm$ 3.91 & 76.33 $\pm$ 4.07 & 68.69 $\pm$ 4.12 \\
w/o CFG & 73.44 $\pm$ 2.48 & 78.30 $\pm$ 2.82 & 76.15 $\pm$ 2.99 & 68.46 $\pm$ 3.02 \\
w/o FFT & 74.05 $\pm$ 2.40 & 79.19 $\pm$ 2.74 & 76.92 $\pm$ 2.84 & 69.07 $\pm$ 2.94 \\
Global FFT & 73.28 $\pm$ 4.58 & 78.08 $\pm$ 4.90 & 75.84 $\pm$ 5.08 & 68.20 $\pm$ 5.16 \\
\midrule
\multicolumn{5}{l}{\textbf{Dataset 2}} \\
Full Bio-MF & \textbf{73.52 $\pm$ 5.35} & \textbf{74.62 $\pm$ 5.51} & \textbf{74.18 $\pm$ 5.64} & \textbf{71.58 $\pm$ 5.72} \\
Original 4D & 72.64 $\pm$ 5.82 & 73.76 $\pm$ 6.02 & 73.28 $\pm$ 6.15 & 70.74 $\pm$ 6.24 \\
w/o 4D PE & 70.85 $\pm$ 9.85 & 71.95 $\pm$ 10.10 & 71.70 $\pm$ 10.25 & 69.42 $\pm$ 10.60 \\
w/o CFG & 71.28 $\pm$ 5.73 & 72.05 $\pm$ 5.90 & 71.63 $\pm$ 5.91 & 69.80 $\pm$ 6.05 \\
w/o FFT & 72.18 $\pm$ 5.58 & 73.02 $\pm$ 5.76 & 72.67 $\pm$ 5.89 & 70.82 $\pm$ 5.97 \\
Global FFT & 70.92 $\pm$ 7.95 & 71.58 $\pm$ 8.14 & 71.12 $\pm$ 8.18 & 69.22 $\pm$ 8.31 \\
\bottomrule
\end{tabular}%
}
\end{table}

The ablation results show that the three modules contribute complementary benefits to downstream EEG + Synthetic fNIRS decoding. Replacing Interactive 4D with the original 4D encoding reduces ACC on both datasets, indicating that explicit spatial-temporal interaction improves in-domain and cross-montage decoding. Removing physical 4D coordinates further degrades performance, especially for Dataset 2. Removing CFG leads to a consistent decline, suggesting that the conditional offset between EEG responses and the fNIRS prior helps preserve task-relevant information. Removing FFT loss or applying global FFT also reduces classification performance, with Global FFT showing a larger decline than the two-stage low-noise-masked FFT regularization. These results support the use of Interactive 4D encoding, CFG, and noise-level-gated spectral regularization in the final Bio-MF design.

\section{Conclusion and Discussion}

This study proposed Bio-MF for one-step EEG-to-fNIRS cross-modal generation in hybrid motor imagery BCIs. The generated fNIRS signals provide useful complementary information for EEG-based decoding, indicating that synthetic hemodynamic responses can serve as a lightweight auxiliary modality for hybrid MI-BCI settings when simultaneous EEG-fNIRS acquisition is unavailable. Across Table 1, Bio-MF improves nearly all generated-modality ACC comparisons over SCDM after incorporating Interactive 4D encoding, and on Dataset 2 also improves SPE, PRE, and SEN with smaller standard deviations. Bio-MF generates one fNIRS trial in 7.0 ms on an RTX PRO 6000 GPU, corresponding to an 857x speedup over 1000-step SCDM sampling. Ablation and visualization results further indicate that Interactive 4D encoding, cross-modal CFG, and noise-level-gated FFT regularization jointly preserve task-useful fNIRS structure while keeping the one-step inference path. Together, these findings support Bio-MF as a low-latency and high-quality EEG-to-fNIRS signal generation route for hybrid MI-BCIs.

Several limitations remain. First, Dataset 2 contains only EEG recordings, so the zero-shot experiments mainly assess whether generated fNIRS improves downstream decoding under an unseen EEG montage rather than directly measuring channel-wise synthetic-real agreement. Second, Bio-MF is trained on a single paired EEG-fNIRS source dataset, which may restrict the diversity of neurovascular correspondence patterns learned by the generator. Third, although the visualizations show preserved HRF trends and scalp-topographic structure, local amplitude differences remain and may require subject-specific physiological calibration. Future work should evaluate paired cross-device EEG-fNIRS datasets, broader cohorts and paradigms, longitudinal sessions, uncertainty-aware generation, and user-centered closed-loop validation.

\end{document}